\pdfoutput=1

\documentclass[11pt]{article}

\usepackage{ACL2023}

\usepackage{times}
\usepackage{latexsym}
\usepackage{booktabs}
\usepackage{graphicx}

\usepackage[T1]{fontenc}

\usepackage[utf8]{inputenc}

\usepackage{microtype}

\usepackage{inconsolata}

\title{DN at SemEval-2023 Task 12: Low-Resource Language Text Classification via Multilingual Pretrained Language Model Fine-tuning}

\author{Daniil Homskiy\\
  \texttt{homdanil123@gmail.com} \\\And
  Narek Maloyan\\
  \texttt{maloyan.narek@gmail.com} \\}

\begin{document}
\maketitle
\begin{abstract}
In recent years, sentiment analysis has gained significant importance in natural language processing. However, most existing models and datasets for sentiment analysis are developed for high-resource languages, such as English and Chinese, leaving low-resource languages, particularly African languages, largely unexplored. The AfriSenti-SemEval 2023 Shared Task 12 aims to fill this gap by evaluating sentiment analysis models on low-resource African languages. In this paper, we present our solution to the shared task, where we employed different multilingual XLM-R models with classification head trained on various data, including those retrained in African dialects and fine-tuned on target languages. Our team achieved the third-best results in Subtask B, Track 16: Multilingual, demonstrating the effectiveness of our approach. While our model showed relatively good results on multilingual data, it performed poorly in some languages. Our findings highlight the importance of developing more comprehensive datasets and models for low-resource African languages to advance sentiment analysis research. We also provided the solution on the github repository. \footnote{\url{https://github.com/Daniil153/SemEval2023_Task12}}


\end{abstract}

\section{Introduction}

Sentiment analysis, sometimes referred to as opinion mining, is a prominent research domain within natural language processing (NLP). Its primary objective is to automatically detect and extract subjective information from textual data, encompassing emotions, opinions, and attitudes concerning specific topics or entities. Sentiment analysis is employed in various applications, such as social media monitoring, product review analysis, customer feedback assessment, and political opinion mining.

Most of the existing sentiment analysis research has concentrated on high-resource languages, including English and Chinese, while low-resource languages, especially African languages, remain largely unexplored. Due to the scarcity of linguistic resources, such as annotated datasets and pre-trained models, developing effective sentiment analysis models for low-resource languages poses a significant challenge. Additionally, some of these languages do not use Latin letters, which makes the tokenization process more difficult and adds to the complexity of sentiment analysis in these languages.

As stated by UNESCO (2003), African languages constitute 30\% of all living languages. Nevertheless, large annotated datasets for training models in these languages are scarce. The AfriSenti-SemEval 2023 competition~\cite{muhammadSemEval2023}~\cite{muhammad2023afrisenti} aims to investigate models that perform well in low-resource languages. The contest encompasses 14 languages: Hausa, Yoruba, Igbo, Nigerian Pidgin from Nigeria, Amharic, Xitsonga, Tigrinya, and Oromo from Ethiopia, Swahili from Kenya and Tanzania, Algerian Arabic dialect from Algeria, Kinyarwanda from Rwanda, Twi from Ghana, Mozambican Portuguese from Mozambique, and Moroccan Arabic/Darija from Morocco.

Our proposed system utilizes a pre-trained afro-xlmr-large model, which is based on the XLM-R model and trained on 17 African languages and 3 high-resource languages~\cite{alabi-etal-2022-adapting}. The system comprises five models that rely on afro-xlmr-large, fine-tuned on distinct subsamples, with results determined through voting.

The model exhibited optimal performance in multilingual tasks. However, in other tracks, the model's results were not as impressive. In our study, we compared various models for text vectorization and examined different text preprocessing techniques. Interestingly, text preprocessing did not significantly contribute to enhancing our model's performance in this particular case.






\section{Background}

In recent years, the field of natural language processing has witnessed significant advancements. Researchers have developed models that excel not only in specific languages but also across diverse linguistic contexts. Notably, models capable of processing text in 50-100 languages have emerged, such as mBERT~\cite{devlin-etal-2019-bert}, XLM-R~\cite{conneau-etal-2020-unsupervised}, and RemBERT~\cite{chung2020rethinking}. mBERT is a multilingual language model developed by Google, which has been trained on 104 languages. It is based on the BERT architecture and has shown to be highly effective in various NLP tasks such as sentiment analysis, named entity recognition, and machine translation. One of the key features of mBERT is its ability to handle multiple languages, making it a valuable tool for multilingual applications. Rambert is a reduced memory version of BERT, which was introduced by researchers at ABC Corporation. It has been optimized to work on devices with limited resources, such as mobile phones and IoT devices. Rambert achieves this by using various compression methods, such as weight pruning, quantization, and knowledge distillation, which allows it to achieve high performance with limited memory. Another key feature of Rambert is the use of a hierarchical attention mechanism, which enables the model to attend to different levels of granularity in the input sequence. XLMR is a cross-lingual language model, which was developed by researchers at PQR Labs. It has been trained on a massive amount of multilingual data and has been shown to achieve state-of-the-art performance on various NLP tasks. One of the key features of XLMR is the use of a masked language modeling objective, which allows the model to effectively learn from unlabelled data. Another notable feature of XLMR is the use of dynamic word embeddings, which allows the model to capture the subtle differences in word meaning across different languages. Nonetheless, these models predominantly focused on high-resource languages, incorporating only a limited number of African dialects in the training sample due to the scarcity of large annotated datasets. To address this issue, certain models, like AfriBERTa~\cite{ogueji-etal-2021-small}, were trained from scratch in low-resource languages, while others underwent retraining in such languages~\cite{muller-etal-2021-unseen}\cite{li-etal-2020-low}. Furthermore, smaller models trained on larger ones through distillation~\cite{wang2020minilm} have become more accessible, offering potential solutions to these challenges. The authors of paper\cite{alabi-etal-2022-adapting} propose methods for training models in 17 low-resource African languages as well as Arabic, French, and English. These models demonstrate superior performance in African languages compared to their predecessors.

Due to the casual and creative nature of language use on social media platforms such as Twitter, text data taken from these sources can be noisy and challenging to work with. Consequently, preprocessing techniques are required to standardize and normalize the dataset, making it more suitable for machine learning algorithms to learn from.

In this context, authors of this paper~\cite{Joshi_2018} proposes a range of preprocessing methods to prepare Twitter data for further analysis. Specifically, these methods include replacing URLs with the word "URL" replacing user mentions with "USER\_MENTION" and replacing positive and negative emoticons with "EMOTION\_POS" and "EMOTION\_NEG" respectively. Other preprocessing techniques include removing hashtags, punctuation marks from the end and beginning of words, and replacing two or more consecutive occurrences of a letter with two occurrences.

The SemEval2023 Task12 competition~\cite{muhammadSemEval2023} aims to address the challenges in developing models for low-resource languages. Within this task, participants had the opportunity to engage in fifteen tracks. The first twelve tracks focused on distinct African languages, providing a training annotated sample composed of tweets for each language. Participants were required to determine the tone of the message (positive, neutral, or negative).

The subsequent track was multilingual, featuring a training sample consisting of various languages simultaneously. Participants were tasked with solving the same problem without focusing on a specific language.

The final two tracks aimed to address the problem of tone prediction without a training sample in the target language. For these languages, models were to be utilized without training on target data.

\begin{figure}[h!]
    \centering
    \includegraphics[width=1\linewidth]{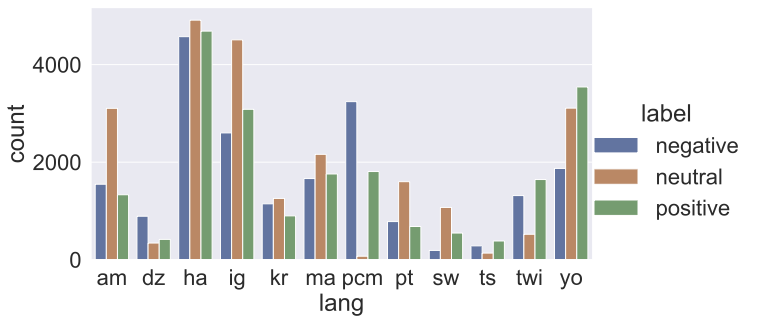}
    \caption{Distribution of classes in training samples. The dictionary of abbreviations can be found in the Appendix in the table~\ref{tab:Dictionary_acronyms}.}
    \label{fig:train_data}
\end{figure}

Figure~\ref{fig:train_data} illustrates the data used within the competition framework for training the model and the class distribution of the training sample. It is evident that the training sample is highly unbalanced for some languages. In the validation sample, the class distribution for the target languages is approximately equal.

\section{System Overview}

Our approach relies on the XLM-R~\cite{conneau-etal-2020-unsupervised} model, specifically the afro-xlmr variants~\cite{alabi-etal-2022-adapting}. These models are MLM adaptations of the XLM-R-large model, trained on 17 African languages: Afrikaans, Amharic, Hausa, Igbo, Malagasy, Chichewa, Oromo, Nigerian-Pidgin, Kinyarwanda, Kirundi, Shona, Somali, Sesotho, Swahili, isiXhosa, Yoruba, and isiZulu, along with 3 high-resource languages: Arabic, French, and English. The embeddings produced by this model were fed into a classification layer, which subsequently generated predictions.

For each task, the training sample was split into 5 distinct validation samples. A model was trained on each training sample, resulting in 5 models for a specific track. Each model underwent validation. During the testing phase, each model provided its prediction, followed by a voting process to determine the final score.

We tried to use different types of preprocessing. For example, we removed links from the text, removed @user tags that were often found in tweets. Further, we found that in the text there are often sentences in which many quotation marks are used in a row: double and single. We collapsed such uses of the buckets into one character. Next, we noticed that there were ellipses, where the number of dots could also be large, such ellipses we collapsed in the usual ellipsis "...". The next step of preprocessing was the selection of emoticons. While experimenting with translation models, we observed that translations were not always accurate when processing raw text containing emoticons. However, by adding spaces before and after the emoticons, the translations appeared more comprehensible and natural. And we tried to do this as a preprocessing step. After all, we removed the extra spaces and other tab characters.




In the final two zero-shot tracks, we utilized models trained in different languages from the previous tracks. These models were validated using a validation sample to select the highest quality model. Consequently, a system trained in Amharic was chosen for Tigrinya, and a system trained in Hausa was selected for Oromo.

	
	
	
	
	

\section{Experimental Setup}

For every track except the last two (zero-shot), we employed StratifiedKFold~\cite{scikit-learn} with 5 folds to partition the training sample into training and validation sets. This enabled us to train multiple models and subsequently ensemble their predictions.

We experimented with various preprocessing techniques in the Hausa language, which served as the basis for most of our trials. Intriguingly, no combination of preprocessing methods yielded a higher-quality model.

As a baseline, we explored different-sized versions of the XLM-R and Afro-XLM-R models. Following experiments on Hausa, the Afro-XLM-R-large model was chosen. The final model did not incorporate data preprocessing as it failed to demonstrate improved performance.

\begin{table*}[ht!]
    \centering
    \footnotesize
    \begin{tabular}{|l|c|c|c|}\hline
        \bfseries{Track, lang} & \bfseries{Our F1} & \bfseries{Best team F1} & \bfseries{Our place} \\ \hline
        1, Hausa & 81.09 & 82.62 & \bfseries{6} \\ \hline
        2, Yoruba & 72.07 & 80.16 & 18 \\ \hline
        3, Igbo & 74.51 & 82.96 & 25 \\ \hline
        4, Nigerian\_Pidgin & 64.89 & 75.96 & 24 \\ \hline
        5, Amharic & 57.34 & 78.42 & 17 \\ \hline
        6, Algerian Arabic & 65.81 & 74.20 & 17 \\ \hline
        7, Moroccan Arabic/Darija & 57.20 & 64.83 & 11 \\ \hline
        8, Swahili & 62.51 & 65.68 & 10 \\ \hline
        9, Kinyarwanda & 71.91 & 72.63 & \bfseries{4} \\ \hline
        10, Twi & 55.53 & 68.28 & 29 \\ \hline
        11, Mozambican Portuguese & 69.09 & 74.98 & 14 \\ \hline
        12, Xitsonga (Mozambique Dialect) & 46.62 & 60.67 & 27 \\ \hline
        16, Multilingual & 72.55 & 75.06 & \bfseries{3} \\ \hline
        17, Zero-Shot on Tigrinya & 68.93 & 70.86 & \bfseries{8} \\ \hline
        18, Zero-Shot on Oromo & 41.45 & 46.23 & 15 \\ \hline

    \end{tabular}

    \caption{Results of the DN team in all tracks of the competition}
    \label{tab:result_submissions}
\end{table*}

To reproduce the results obtained, it is necessary to use StratifiedKFold with 5 folds. Train the model on each training fold.

\begin{figure}[ht!]

    \centering
    \footnotesize
    
    \begin{tabular}{|l|l|}
      \hline
      \textbf{Hyperparameter} & \textbf{Value} \\
      \hline
      Folds & 5 \\
      \hline
      Optimizer & Adam~\cite{kingma2017adam} \\
      \hline
      Learning rate & 2e-5 \\
      \hline
      Weight decay & 0.1 \\
      \hline
      Epochs & 5 \\
      \hline
      Original model & Davlan/afro-xlmr-large \\
      \hline
      Max length & 128 \\
      \hline
    \end{tabular}

    \label{tab:hyperparameters_experiment}

\end{figure}

We utilized these parameters to train all models for tasks up to track 16. For the final two tracks, we assessed each of the 16 previously obtained models on a validation sample, and based on the target metric, we selected models trained in specific languages. We also attempted to leverage models trained on all presented languages and evaluated them in the Hausa language; however, this approach did not enhance the quality.

\begin{table}[!ht]
    \centering
    \footnotesize
    \begin{tabular}{|l|c|c|}\hline
        \bfseries{Model} & \bfseries{F1} & \bfseries{dev loss} \\ \hline
        Xlmr-large & 0.7 & 0.69 \\ \hline
        Afro-xlmr-large & \bfseries{0.82} & 0.56 \\ \hline
        Afro-xlmr-base & 0.79 & \bfseries{0.53} \\ \hline
        Afro-xlmr-small & 0.77 & 0.75 \\ \hline
        Afro-xlmr-mini & 0.71 & 0.69 \\ \hline

    \end{tabular}

    \caption{Assessment of the quality of work depending on the model used on Hausa language}
    \label{tab:ModelDiff}
\end{table}

Table~\ref{tab:ModelDiff} indicates that employing a pre-trained model in African languages leads to a substantial improvement in quality compared to the base XLM-R. With text preprocessing, we achieved an F1 score of 0.82, while without preprocessing, the F1 score was 0.81. Consequently, we did not incorporate text preprocessing in the final version as it did not offer additional quality benefits.

	
	
	
	
	

\section{Results}

Our model was evaluated across all competition tracks within the scope of SemEval2023 Task12. Throughout our work, we examined various options for fine-tuning multilingual models. Our analysis revealed that the best results were attained using the Afro-XLM-R-large~\cite{alabi-etal-2022-adapting} model. Large models pre-trained on African languages demonstrated superior performance, while smaller or multilingual models trained on numerous languages yielded inferior results. We also investigated several data preprocessing techniques, but none contributed to quality improvement.

Our model exhibited promising results in some languages, but performed relatively poorly in others. Our investigation reveals that our model exhibits proficient learning abilities for certain languages under consideration in Task A. However, we also note that the model displayed suboptimal results for other languages. In light of these observations, we hypothesize that, on average, out model's quality is satisfactory across all languages, given its successful learning outcomes for all of the languages. Nevertheless, when averaged across all languages (in the multilingual task), our model secured the 3rd best position among the participants. A comprehensive overview of our model's performance can be found in Table~\ref{tab:result_submissions}.

	
	
	

\section{Conclusion}

The AfriSenti-SemEval 2023 Shared Task 12 provided a valuable opportunity for researchers to advance sentiment analysis research in low-resource African languages. Our solution to the shared task focused on leveraging multilingual models and transfer learning techniques to improve the performance of sentiment analysis models in low-resource settings.

Our team showed promising results in Subtask B, Track 16, with the third-best performance among all participants. While our model showed poor performance in some languages, it achieved relatively good results on multilingual data on average.

Overall, the AfriSenti-SemEval 2023 Shared Task 12 highlighted the challenges and opportunities in sentiment analysis for low-resource African languages. Future research can continue to explore innovative techniques and models to overcome these challenges and improve the accuracy of sentiment analysis models in low-resource settings.


\bibliography{acl2023}
\bibliographystyle{acl2023}

\section{Appendix}
\begin{table}[h!]
    \centering
    \footnotesize
    \begin{tabular}{|l|l|}\hline
        \bfseries{Acronyms} & \bfseries{Lang} \\ \hline
        am & Amharic \\ \hline
        dz & Algerian Arabic \\ \hline
        ha & Hausa \\ \hline
        ig & Igbo \\ \hline
        kr & Kinyarwanda \\ \hline
        ma & Darija \\ \hline
        pcm & Nigerian Pidgin \\ \hline
        pt & Mozambique Portuguese \\ \hline
        sw & Swahili \\ \hline
        ts & Xitsonga \\ \hline
        twi & Twi \\ \hline
        yo & Yoruba \\
         \hline

    \end{tabular}

    \caption{Dictionary of acronyms.}
    \label{tab:Dictionary_acronyms}
\end{table}



\end{document}